\pdfoutput=1
\documentclass{article}

\PassOptionsToPackage{numbers, compress}{natbib}

\usepackage[main, final]{neurips_2026}

\usepackage[utf8]{inputenc}
\usepackage[T1]{fontenc}
\usepackage{hyperref}
\usepackage{url}
\usepackage{booktabs}
\usepackage{amsfonts}
\usepackage{nicefrac}
\usepackage{microtype}
\usepackage{siunitx}
\usepackage[table]{xcolor}
\usepackage{multirow}
\usepackage{amsmath}
\usepackage{cleveref}
\usepackage{graphicx}
\usepackage{pifont}
\usepackage{wrapfig}
\usepackage{capt-of}
\usepackage{needspace}
\newcommand{\cmark}{\textcolor{green!60!black}{\ding{51}}}
\newcommand{\xmark}{\textcolor{red!60}{\ding{55}}}

\title{TESLA: Native 4D Gaussian Splatting Generation with Temporally Structured Latents}

\author{%
  \textbf{Weiqi Zhang}$^{1*}$\quad \textbf{Junsheng Zhou}$^{1*\dagger}$\quad \textbf{Juntong Fang}$^{1*}$ \\
  \textbf{Kanle Shi}$^{2}$\quad \textbf{Shenkun Xu}$^{2}$\quad \textbf{Yu-Shen Liu}$^{1\dagger}$ \\[0.1cm]
  {\normalfont\small $^{1}$School of Software, Tsinghua University, Beijing, China\quad $^{2}$Kuaishou Technology, Beijing, China} \\
  {\ttfamily\small \{zwq23, zhou-js24, fangjt21\}@mails.tsinghua.edu.cn} \\
  {\ttfamily\small \{shikanle,xushenkun\}@kuaishou.com\quad liuyushen@tsinghua.edu.cn}
}

\begin{document}

\maketitle

\begingroup
\renewcommand{\thefootnote}{}
\footnotetext{$^{*}$Equal contribution. $^{\dagger}$Corresponding authors.}
\endgroup

\begin{figure}[h]
    \vspace{-0.2cm}
    \centering
    \includegraphics[width=0.84\textwidth]{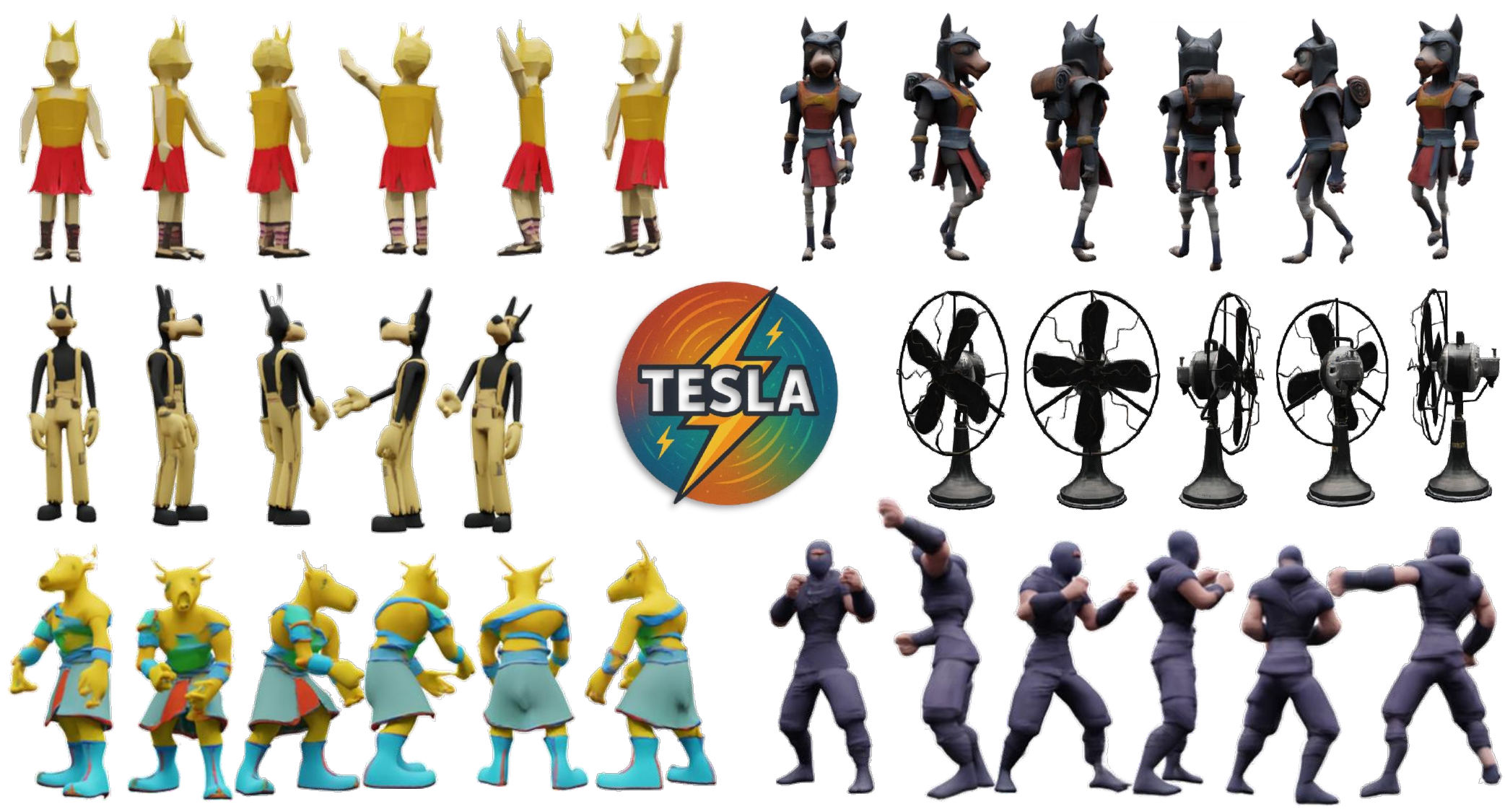}
    \vspace{-0.2cm}
    \caption{Given a reference image and text description, TESLA generates native 4D Gaussian Splatting with realistic and temporally coherent animations.}
    \label{fig:teaser}
    \vspace{-0.4cm}
\end{figure}

\begin{abstract}
Diffusion models have achieved remarkable success in image, video, and 3D content creation.
However, extending these successes to native 4D model generation remains challenging due to the intricate coupling between temporal geometry and appearance dynamics.
Existing approaches typically generate 4D content by synthesizing videos from novel viewpoints, yet they often suffer from poor spatio-temporal consistency across views and time.
To address these limitations, we introduce TESLA, a novel feedforward approach for generating native 4D Gaussian Splatting (4DGS) representations.
Our key insight is to decouple temporal geometry and appearance generation through a temporally structured latent representation. This decomposition allows us to first establish coarse temporal geometry that captures fundamental structural dynamics, and then synthesize fine-grained appearance details conditioned on this geometric scaffold.
To bridge this decoupled representation to the final output, we design a native 4DGS decoder that directly transforms our temporal features into native 4DGS representations, enabling flexible temporal rendering across arbitrary viewpoints and timesteps.
Trained on a large collection of curated animated sequences, we extensively evaluate TESLA on the 4DGS generation task, showing superior performance over existing baselines. Project Page: \url{https://weiqi-zhang.github.io/TESLA}.

\end{abstract}

\section{Introduction}

Dynamic 4D content generation represents a cornerstone technology for next generation digital experiences across AR/VR, gaming, and digital media, enabling the creation of temporally coherent dynamic objects that evolve naturally over time. While recent advances in diffusion models have demonstrated remarkable progress in static 3D object generation~\cite{hunyuan3d22025tencent,he2025triposf,xiang2024structured,zhou2024diffgs,zhou2024udiff}, extending these achievements to the temporal domain remains fundamentally challenging.
The key difficulties lie in modeling spatio-temporal dynamics, maintaining inter-frame geometric and appearance consistency, and managing the exponential growth in computational complexity introduced by temporal dimensions.

Various representation forms exist for modeling 4D dynamic scenes, such as animated mesh and deformable 3DGS. More recently, 4D Gaussian Splatting~\cite{yang2023gs4d} has emerged as a promising 4D representation that offers unique advantages over existing approaches. While mesh-based representations are constrained by fixed topology and skeletal structures, 4D Gaussian Splatting naturally handles complex deformations, appearance changes, and topological variations through independent Gaussian evolution.

Current 4DGS generation approaches fall into two primary categories. The first employs gradient optimization using powerful 2D diffusion models as supervisory signals~\cite{poole2022dreamfusion}. These methods~\cite{bah20244dfy,ren2023dreamgaussian4d,wu2024sc4d,zhao2023animate124} iteratively refine 4D representations by distilling knowledge from pre-trained generative models, but suffer from computational intensity and optimization instabilities that cause temporal flickering and geometric inconsistencies. The second category employs reconstruction pipelines that generate multi-view videos before reconstructing 4D Gaussian representations. These approaches~\cite{liang2024diffusion4d,jin2025diffuman4d,wu2025cat4d} suffer from viewpoint inconsistencies and temporal artifacts that propagate through reconstruction. More fundamentally, native 4DGS generation faces inherent challenges from high-dimensional spatio-temporal data where each Gaussian primitive must maintain temporal coherence across the entire sequence.

To overcome these fundamental challenges, we introduce \textbf{TESLA}, a novel feedforward approach that natively generates 4DGS representations without relying on intermediate reconstruction or optimization. The key insight of TESLA lies in a temporally structured latent representation that decouples geometric evolution from appearance synthesis. By first establishing coarse temporal geometry that captures fundamental structural changes and then conditioning fine-grained appearance synthesis on this geometric scaffold, TESLA effectively reduces the complexity of high-dimensional spatio-temporal modeling.

In addition to the decoupled paradigm, TESLA introduces a native 4DGS decoder that directly predicts Gaussian parameters from temporally structured latent representations, thereby eliminating intermediate reconstruction stages. This native generation scheme enables arbitrary temporal sampling during rendering, free from the constraints of discrete video frames. Moreover, the decoder integrates temporal attention mechanisms to enhance temporal coherence and appearance consistency across the generated sequence.

We curate a comprehensive dataset comprising \textbf{45,000} animated sequences collected from Deformthings4D~\cite{li20214dcomplete}, Objaverse~\cite{objaverse} and Objaverse-XL~\cite{objaverseXL}, totaling over \textbf{1.08 million} frames across diverse object categories. Each frame is rendered from multiple viewpoints to provide rich multi-view supervision. The entire curation and rendering process requires approximately \textbf{50,000} GPU hours, resulting in a large-scale foundational dataset that we expect to benefit future research in 4D content generation. Trained on this dataset, TESLA achieves significant improvements over existing methods, with extensive evaluation demonstrating superior animation quality and computational efficiency compared to prior approaches.

Our primary contributions include:
\begin{itemize}
\item We introduce TESLA, a novel feedforward framework for generating 4D content as native 4D Gaussian Splatting representation. TESLA is designed based on temporally structured latents that effectively decouples temporal geometry and appearance dynamics, simplifying the inherent complexity of 4D data modeling.

\item We develop a native 4D Gaussian Splatting decoder that directly predicts Gaussian parameters from temporal latents, enabling flexible and efficient 4DGS generation.

\item We contribute a meticulously curated dataset of \textbf{45,000} animated sequences requiring about \textbf{50,000} GPU hours of rendering, with comprehensive evaluations demonstrating TESLA's superior performance in generating high-quality 4DGS representations.
\end{itemize}

\section{Related Work}

\subsection{Native 3D Generation}
The field of 3D content generation has evolved significantly, transitioning from direct manipulation of raw 3D data to sophisticated latent-based approaches. Initial methodologies operated directly on explicit 3D representations including point clouds~\cite{nichol2022point}, volumetric grids~\cite{gupta20233dgen}, and implicit neural representations such as Neural Radiance Fields~\cite{mildenhall2020nerf,cao2023large,gao2022get3d}. While these direct approaches were conceptually straightforward, they struggled with scalability and poor generalization due to limited datasets and computational constraints.

This field has increasingly adopted latent-based denoising approaches for 3D content creation~\cite{wu2025direct3ds2gigascale3dgeneration,li2024craftsman,zhao2023michelangelo,li2025triposg}, utilizing variational autoencoders to encode high-dimensional geometric information into compact latent representations that facilitate efficient generative modeling. Early works~\cite{shape2vec,Chen_2025_Dora} pioneered vector-based latent encoding strategies for geometric data compression and reconstruction. CLAY~\cite{zhang2024clay} established the viability of merging VAE architectures with transformer-driven diffusion processes, achieving competitive performance on large-scale datasets with flexible input modalities. TripoSG~\cite{li2025triposg} advanced the field by implementing rectified flow techniques~\cite{liu2022flow} alongside mixture-of-experts designs to accelerate generation workflows while maintaining quality. Trellis~\cite{xiang2024structured} introduced hierarchical sparse encoding schemes coupled with adaptive decoding modules to enhance flow-model training efficiency and representation quality. While these latent-based 3D generation methods, together with methods that generate 3D Gaussians~\cite{zhou2024diffgs,zhang2025gap,zhang2026gaussiangrow} and unsigned distance fields~\cite{zhou2024udiff,zhou2026udfstudio}, have achieved remarkable success in creating static geometric content, extending these achievements to temporal dynamics and 4D generation presents fundamental challenges. The addition of temporal dimension introduces complexities in modeling coherent motion patterns, maintaining inter-frame consistency, and handling the exponential growth in data dimensionality across time-varying scenes. Reconstruction methods built on Gaussian Splatting~\cite{zhang2025materialrefgs,li2025gaussianudf} or signed distance functions~\cite{noda2024multipull,noda2025bijective} likewise target static scenes and shapes.

\subsection{4D Generation}
4D content generation represents a frontier challenge in generative modeling, requiring the synthesis of both geometric structure and temporal dynamics within a unified framework. Early approaches have predominantly employed distillation techniques that leverage knowledge from established 2D, 3D, or video generation models~\cite{bah20244dfy,jiang2024consistentd,ling2024align,ren2023dreamgaussian4d,zhao2023animate124} to guide spatio-temporal synthesis. These distillation-based methods, however, impose significant computational burdens and necessitate lengthy optimization procedures for individual scenes, often resulting in temporal artifacts and spatial inconsistencies that degrade generation quality. More recent work has investigated fine-tuning strategies applied to existing 3D and video generation architectures using available 4D datasets, enabling the production of multi-view dynamic sequences~\cite{liang2024diffusion4d,jin2025diffuman4d,wu2025cat4d,yin2025splat4d,pan2025diff4splat,jiang2025geo4d,yu2025trajectorycrafter} with reduced computational requirements and improved temporal stability. Nevertheless, these approaches remain limited by their reliance on per-scene reconstruction workflows and exhibit spatial discrepancies due to the scarcity of high-quality 4D training data. Feed-forward 4D reconstruction models~\cite{ren2024l4gm,zhang2026gamma,fang2026more,chen2026uniquery4r} avoid per-scene optimization, but they recover dynamic content from an input video rather than generating it. Moving-region segmentation under camera motion~\cite{jin2026morse} likewise analyzes motion already present in a video. Recently, AnimateAnyMesh~\cite{wu2025animateanymesh} pioneered feed-forward mesh animation through trajectory decomposition and topology-aware encoding. GVDiffusion~\cite{zhang2025gaussian} proposed a Gaussian Variation Field approach that efficiently compresses 4D mesh animations into compact latent spaces and enables video-to-4D generation through diffusion modeling. However, these methods primarily require intermediate conversions, and cannot directly generate native 4D Gaussian Splatting with flexible rendering. Concurrently, AEON~\cite{zhang2026aeon} decodes video diffusion latents into 4D Gaussians for scene-level world modeling, whereas TESLA generates object-level animations natively from temporally structured latents.

\section{Method}
\begin{figure*}[t]
    \centering
    \includegraphics[width=\linewidth]{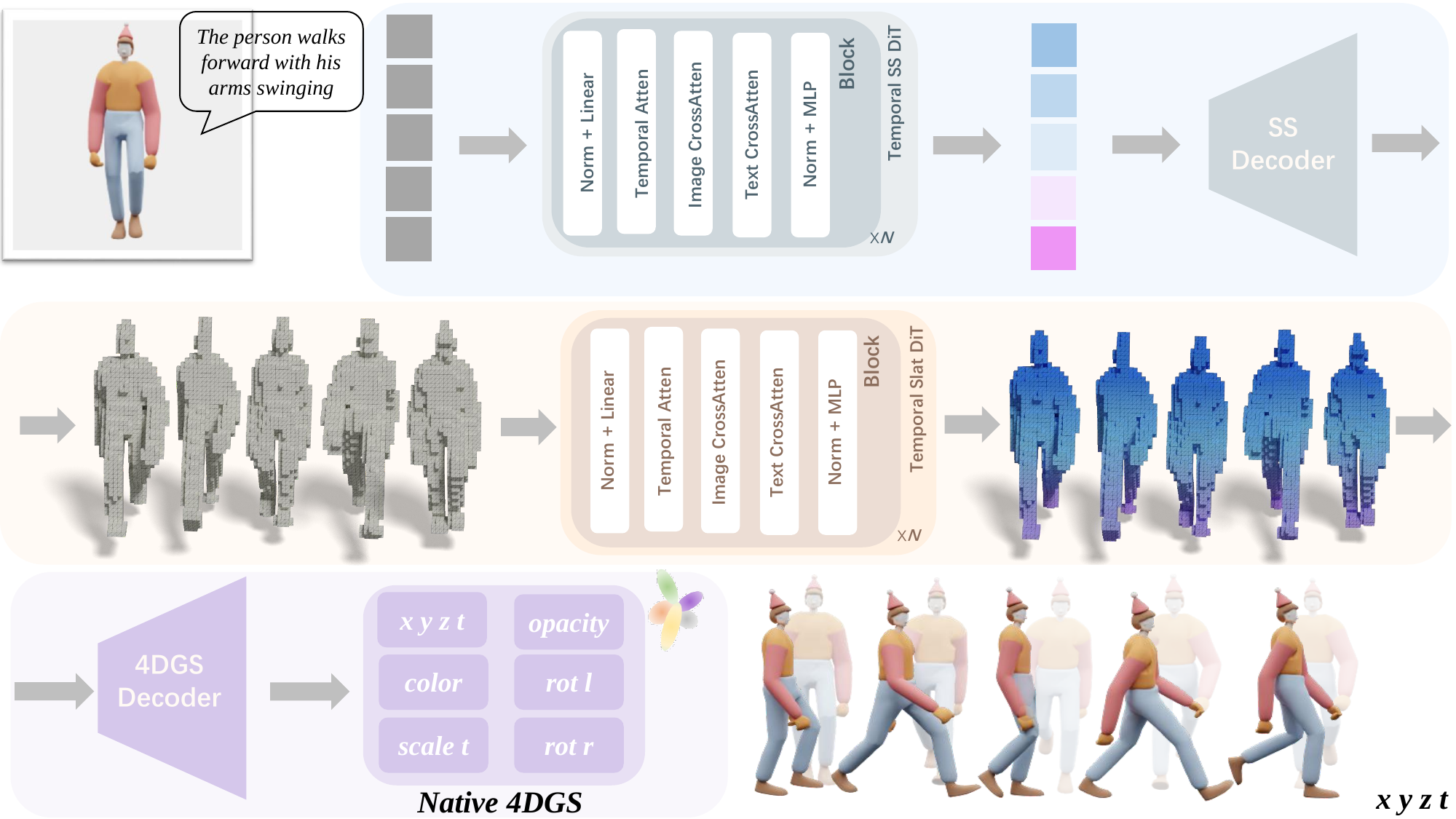}
    \vspace{-0.6cm}
    \caption{Overview of TESLA for native 4D Gaussian Splatting generation. Unlike methods that require intermediate video reconstruction or per-scene optimization, TESLA directly generates native 4DGS in a feedforward manner. Given a reference image and text prompt, the framework consists of three stages: (1) The inputs are encoded into multi-modal conditioning features. (2) A Temporal Slat DiT with parallel attention architecture generates structured temporal latents, where Temporal SS DiT first produces sparse geometric structures, followed by dense appearance synthesis. (3) The 4DGS Decoder directly transforms temporal latents into native 4D Gaussian parameters, enabling flexible temporal rendering without being constrained to discrete video frames. The parallel attention mechanism processes image and text conditions simultaneously to ensure effective multi-modal guidance while maintaining temporal coherence.}
    \label{fig:overview}
    \vspace{-0.3cm}
\end{figure*}

We introduce TESLA, a novel framework that generates native 4D Gaussian Splatting representations through a feedforward approach that decouples temporal geometry and appearance dynamics. Given a reference image $I$ and descriptive text prompt $P$, our goal is to generate temporally coherent 4DGS representations $G$. The overview of TESLA is shown in Fig.~\ref{fig:overview}. We begin by previewing 4D Gaussian Splatting and introducing data preparation in Sec.~\ref{sec:3.1}. In Sec.\ref{sec:3.2}, we present the encoding and decoding framework that establishes a structured latent and enables native 4DGS generation. We further detail the temporal generation approach that synthesizes temporally structured latents through diffusion transformers~\cite{Peebles2022DiT} conditioned on multi-modal inputs in Sec.~\ref{sec:3.3}.

\subsection{Preliminary Knowledge}
\label{sec:3.1}
\noindent\textbf{Preview 4D Gaussian Splatting}. 4D Gaussian Splatting (4DGS)~\cite{yang2023gs4d} extends 3D Gaussian Splatting to dynamic scenes by representing the underlying spatio-temporal 4D volume as a collection of 4D Gaussians that capture both temporal geometry evolution and time-varying view-dependent appearances. For a 4DGS representation $G = {\{gi\}}^N_{i=1}$, each Gaussian $g_i$ models the spatio-temporal distribution through a 4D covariance matrix $\Sigma_i \in R^{4\times4}$ and center $\mu_i \in R^4$, formulated as:
\begin{equation}
 g_i(x,y,z,t) = \exp\!\left[
  -\tfrac{1}{2}\bigl(\xi - \mu_i\bigr)^{\mathrm T}
  \Sigma_i^{-1}
  \bigl(\xi - \mu_i\bigr)
\right],
\end{equation}
where $\xi = [x, y, z, t]^{\top}$ denotes the 4D spatio-temporal coordinates. Following the factorization strategy of 3DGS, the covariance matrix is decomposed as
\begin{equation}
\Sigma_i = R_i\, S_i S_i^{\top} R_i^{\top},
S_i = \operatorname{diag}(s_x, s_y, s_z, s_t),
\end{equation}
where $R_i$ represents a 4D rotation matrix and $S_i$ encodes anisotropic scaling across spatial and temporal dimensions. Crucially, the 4D rotation $R_i$ is parameterized using left and right isoclinic rotations via quaternion pairs $(\mathbf{q}_{\ell}, \mathbf{q}_{r})$, which enables the Gaussian to undergo arbitrary rotations in 4D space and naturally capture complex temporal dynamics.

During rendering at a specific time $t$, each 4D Gaussian undergoes probabilistic decomposition into a conditional 3D Gaussian $p(x,y,z \mid t)$ that governs spatial geometry and a marginal 1D temporal Gaussian $p(t)$ that provides time-dependent weighting. This decomposition allows the rendering pipeline to leverage efficient 3D splatting while incorporating temporal coherence. For appearance modeling, each Gaussian maintains an opacity value $\alpha_i \in \mathbb{R}$ and employs time-evolving view-dependent color $\mathbf{c}_i(\mathbf{d}, t) \in \mathbb{R}^3$, where $\mathbf{d}$ denotes the viewing direction.

\subsection{Encoding and Decoding for 4DGS}
\label{sec:3.2}
\noindent\textbf{Temporal Latent Representation}. 4D Gaussian Splatting generation faces inherent complexity due to the intricate coupling between geometric motion and appearance variations across time. The established paradigms in image-to-video generation fundamentally rely on temporalizing spatial representations, which proves inadequate for handling such complex spatiotemporal dynamics. Inspired by the success of sparse structural approaches in 3D generation~\cite{xiang2024structured}, we extend the structured latent paradigm to the temporal domain to effectively address this limitation. Our approach introduces a temporal structured latent representation that organizes 4D information across both spatial and temporal dimensions. We construct a temporal sequence of structured latents $S = \{S_t\}_{t=1}^{T}$, where each $S_t = \{(z_{{i,t}}, p_{{i,t}})\}_{i=1}^{L_t}$ represents the structured latent at time step $t$. Here, $p_{{i, t}} \in {0,1,\ldots,(N-1)}^3$ denotes the positional indices of active voxels within the 3D grid occupied by the object at time $t$, capturing geometric structure evolution, while $z_{{i,t}}\in\mathbb{R}^{C}$ represents the local latent features attached to these voxels, encoding detailed appearance information that varies across space and time.

\noindent\textbf{Decoding For 4DGS}. Given the temporal structured latent representation $S = {\{S_t\}}^T_{t=1}$, we design a native 4DGS decoder to transform these temporally structured latents directly into native 4D Gaussian primitives without intermediate reconstruction stages. Built upon Transformer~\cite{vaswani2017attention}, the decoder employs temporal self-attention layers to capture motion dynamics and ensure temporal consistency across frames while modeling the evolution of Gaussian attributes including 4D positions, scales, rotations, colors, and opacities. Since the temporal latents inherently encode sequential information across time steps, our decoder preserves the temporal information of each Gaussian while additionally learning a temporal offset parameter $\Delta t_i^{(t)}$ that captures the dynamic motion patterns of individual Gaussians. The decoder output can be formulated as:
\begin{align}
\mathcal{G} &= {\{G_t\}}_{t=1}^T = \text{Decoder}(S), \\
G_t &= {(\mu_i^{(t)}, \mathbf{s}_i^{(t)}, \mathbf{q}_{\ell,i}^{(t)},
\mathbf{q}_{r,i}^{(t)}, \mathbf{c}_i^{(t)}, \alpha_i^{(t)}, \Delta t_i^{(t)})}_{i=1}^{N}.
\end{align}
where $\mu_i^{(t)} \in \mathbb{R}^4$ denotes the 4D center position $(x, y, z, t)$, $\mathbf{s}_i^{(t)} \in \mathbb{R}^4$ represents the 4D scale parameters $(s_x, s_y, s_z, s_t)$, $\mathbf{q}_{\ell,i}^{(t)}, \mathbf{q}_{r,i}^{(t)} \in \mathbb{R}^4$ are the left and right quaternion rotation pairs, $\mathbf{c}_i^{(t)} \in \mathbb{R}^3$ indicates the color, $\alpha_i^{(t)} \in \mathbb{R}$ represents the opacity, and $\Delta t_i^{(t)} \in \mathbb{R}$ denotes the learned temporal offset that enables flexible motion modeling for the $i$-th Gaussian at time $t$. Each temporal latent $S_t$ is processed through multiple transformer blocks that incorporate positional embeddings to maintain spatial-temporal correspondence, enabling the direct synthesis of coherent 4D Gaussian Splatting sequences with enhanced motion representation capabilities.

\noindent\textbf{Training Loss}. The design of training objective focuses on two fundamental challenges in 4D generation: achieving high-quality visual synthesis and maintaining motion consistency. We design a multi-component loss function that systematically tackles these objectives.

For visual quality, we employ a comprehensive reconstruction loss that captures multiple aspects of image fidelity:
\begin{equation}
\mathcal{L}_{\text{recon}} = \mathcal{L}_1 + 0.2(1 - \text{SSIM}) + 0.2\mathcal{L}_{\text{LPIPS}}.
\end{equation}
For motion consistency, we introduce a motion regularization loss that directly constrains Gaussian velocity to prevent erratic movements:
\begin{equation}
\mathcal{L}_{\text{motion}} = \frac{1}{N} \sum_{i=1}^{N} |v_i|_2.
\end{equation}
Additionally, we include two auxiliary regularization terms to maintain stable Gaussian properties. The volume regularization loss is defined as:
\begin{equation}
\mathcal{L}_{\text{vol}} = \frac{1}{LK} \sum_{i=1}^{L} \sum_{k=1}^{K} s_k^i,
\label{eq:vol}
\end{equation}
which prevents excessive Gaussian concentration. The opacity regularization loss is:
\begin{equation}
\mathcal{L}_{\alpha} = \frac{1}{LK} \sum_{i=1}^{L} \sum_{k=1}^{K} (1 - \alpha_k^i)^2,
\label{eq:alpha}
\end{equation}
which constrains Gaussians from being excessively transparent. The complete training objective is:
\begin{equation}
\mathcal{L}_{\text{total}} = \mathcal{L}_{\text{recon}} + \lambda_{\text{motion}} \mathcal{L}_{\text{motion}} + \lambda_{\text{vol}} \mathcal{L}_{\text{vol}} + \lambda_{\alpha} \mathcal{L}_{\alpha}.
\end{equation}

\subsection{Temporal Latent Generation}
\label{sec:3.3}
\noindent\textbf{Rectified Flow Model}. To generate temporal slat representations, we employ a Rectified Flow model~\cite{liu2022flow} as our core generative framework. Unlike traditional diffusion models such as DDPM~\cite{ho2020denoising} that rely on complex noise schedules, Rectified Flow learns straight-line trajectories from noise to data distributions, offering superior efficiency and training stability.

Given target temporal latent representations $S = {\{S_t\}}_{t=1}^T$, our objective is to train a neural network $v_\theta(x, t, I, P)$ that predicts the velocity field for transforming noise $x_0 \sim \mathcal{N}(0, I)$ to target data $x_1 = S$, where $I$ represents the reference image and $P$ denotes the text prompt describing the desired animation. The Rectified Flow training objective is formulated as:
\begin{equation}
    \mathcal{L}_{RF} = \mathbb{E}_{t \sim \mathcal{U}[0,1], x_0, x_1} \|v_\theta(x_t, t, I, P) - (x_1 - x_0)\|^2,
\end{equation}
where $x_t = tx_1 + (1-t)x_0$ represents linear interpolation along the optimal transport path. This formulation encourages direct mapping between noise and structured temporal representations conditioned on both visual and textual information, significantly reducing computational overhead during inference while ensuring temporal coherence guided by the reference image appearance and text-described motion dynamics.

\noindent\textbf{Model Architecture}. Our temporal latent generation model builds upon the Diffusion Transformer architecture~\cite{Peebles2022DiT}, adapted for dual-conditioned temporal synthesis. The model generates temporal latent embeddings conditioned on reference images and text prompts through three key architectural innovations designed for effective 4D generation.

\textbf{\textit{Temporal Attention Mechanism}} is designed to maintain temporal coherence in 4DGS generation by introducing dedicated temporal self-attention layers that operate across the time dimension. By treating spatial dimensions as independent sequences, this mechanism focuses specifically on temporal relationships and computes attention weights between different timesteps. This design enables the model to capture long-range temporal dependencies and ensure smooth motion transitions throughout the generation process.

\begin{wrapfigure}{r}{0.32\linewidth}
    \centering
    \vspace{-0.3cm}
    \includegraphics[width=\linewidth]{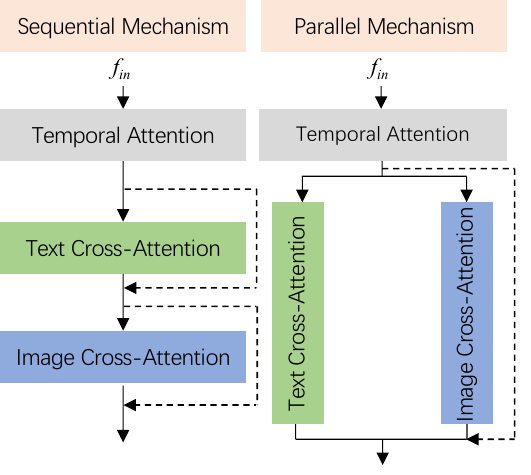}
    \vspace{-0.6cm}
    \caption{Attention mechanisms for multi-modal conditioning.}
    \vspace{-0.4cm}
    \label{fig:paralel}
\end{wrapfigure}

\textbf{\textit{Parallel Attention Architecture.}}
To enable effective multi-modal conditioning, a parallel attention mechanism is adopted, in which image and text guidance are processed simultaneously through separate cross-attention branches. The image cross-attention branch preserves appearance consistency based on reference visuals, while the text cross-attention branch captures motion semantics conveyed by the textual descriptions.

In contrast to sequential conditioning approaches, which process modalities in order and may incur information bottlenecks due to inter-modal dependencies, this architecture adopts a parallel structure. Temporal attention is applied to model motion dynamics:
\begin{equation}
f_{\text{temp}} = \operatorname{TemporalAttn}(f_{\text{in}}) + f_{\text{in}},
\end{equation}

Subsequently, both text and image conditioning are applied in parallel:
\begin{equation}
f_{\text{out}} = \operatorname{TextCrossAttn}(f_{\text{temp}}, f_{\text{text}}) + \operatorname{ImageCrossAttn}(f_{\text{temp}}, f_{\text{ref}}) + f_{\text{temp}},
\end{equation}

where $f_{\text{ref}}$ and $f_{\text{text}}$ denote the reference image features and text embeddings, respectively. This parallel formulation allows each modality to attend independently to the temporally enriched features, thereby avoiding sequential dependencies and facilitating more effective multi-modal fusion for temporal generation. We further validate this parallel design against a sequential conditioning baseline in Tab.~\ref{tab:architecture_ablation}.

\textbf{\textit{Spatio-Temporal Position Encoding }} employs a hybrid $xyzt$ position embedding scheme that captures both spatial coordinates within each timestep and temporal progression across the sequence. This enables the attention layers to understand spatio-temporal correlations, which is essential for coherent 4D generation. The full two-stage inference procedure is detailed in Appendix~\ref{sec:appendix_inference}.

\section{Experiments}

\noindent\textbf{Implementation Details}. We train TESLA using NVIDIA A800 GPUs across three stages. The 4DGS decoder is trained on 8 GPUs for 5 days, while the temporal SS flow model and slat flow model are trained on 32 GPUs for 8 days and 10 days respectively. For data preparation, we employ Qwen2.5-VL~\cite{Qwen2.5-VL} to generate detailed captions for each animated sequence. This automatic captioning process ensures comprehensive text-image-motion alignment across our dataset of 45,000 sequences. For the loss function, we set $\lambda_{\text{motion}} = 0.1$, $\lambda_{\text{vol} }= 10000.0$, and $\lambda_{\alpha} = 0.1$. The temporally structured latent representation uses a grid resolution of $R = 64$, generating sequences of $T = 24$ frames during training. Each frame is rendered from 36 or 48 uniformly sampled viewpoints to provide multi-view supervision.

\begin{table}[t]
  \centering
  \caption{Quantitative comparison on animation generation. Objective metrics are from VBench; user-study scores (1--5) are averaged over 20 participants on motion--text alignment (M.Match), realism (M.Real.), and geometric consistency (G.Cons.).}
  \vspace{-0.2cm}
  \resizebox{\linewidth}{!}{%
  \begin{tabular}{l|cccc|ccc}
    \toprule
    & \multicolumn{4}{c|}{Objective Metrics} & \multicolumn{3}{c}{User Study (1--5)} \\
    \cmidrule(lr){2-5} \cmidrule(lr){6-8}
    Method & I2V~$\uparrow$ & M.Sm.~$\uparrow$ & Sub.Cons.~$\uparrow$ & Aest.Q~$\uparrow$ & M.Match~$\uparrow$ & M.Real.~$\uparrow$ & G.Cons.~$\uparrow$ \\
    \midrule
    Animate3D~\cite{jiang2024animate3d} & 0.663 & 0.990 & 0.727 & 0.420 & 1.850 & 1.717 & 1.917 \\
    AnimateAnyMesh~\cite{wu2025animateanymesh} & 0.850 & 0.993 & 0.911 & 0.532 & 3.217 & 3.483 & 3.350 \\
    Ours & \textbf{0.861} & \textbf{0.994} & \textbf{0.941} & \textbf{0.533} & \textbf{3.850} & \textbf{3.775} & \textbf{3.867} \\
    \bottomrule
  \end{tabular}}
  \label{tab:animation}
  \vspace{-0.5cm}
\end{table}

\subsection{Animation Generation}
\label{sec.4.1}
\noindent\textbf{Evaluation and Metrics}.  Following the evaluation protocol established in Animate3D~\cite{jiang2024animate3d} and AnimateAnyMesh~\cite{wu2025animateanymesh}, we conduct both objective measurements and subjective assessments. For objective evaluation, we utilize three perceptual metrics from the VBench suite~\cite{zhang2024evaluationagent}, consistent with the assessment framework in Animate3D~\cite{jiang2024animate3d}: I2V Subject Similarity, Aesthetic Quality, Motion Smoothness and Subject Consistency. These metrics are computed on view-aligned renderings by rendering the generated 4D Gaussian Splatting from predetermined camera viewpoints.

To supplement our quantitative evaluation, we conducted a perceptual study involving 20 participants. Each participant assessed the generated animations using a five-point rating system, with higher scores reflecting superior quality. The evaluation focused on three key aspects: how well the motion matched the textual description, the realism of the movements, and the consistency of geometric structure throughout the animation.

\begin{figure}[t]
    \centering
    \includegraphics[width=\linewidth]{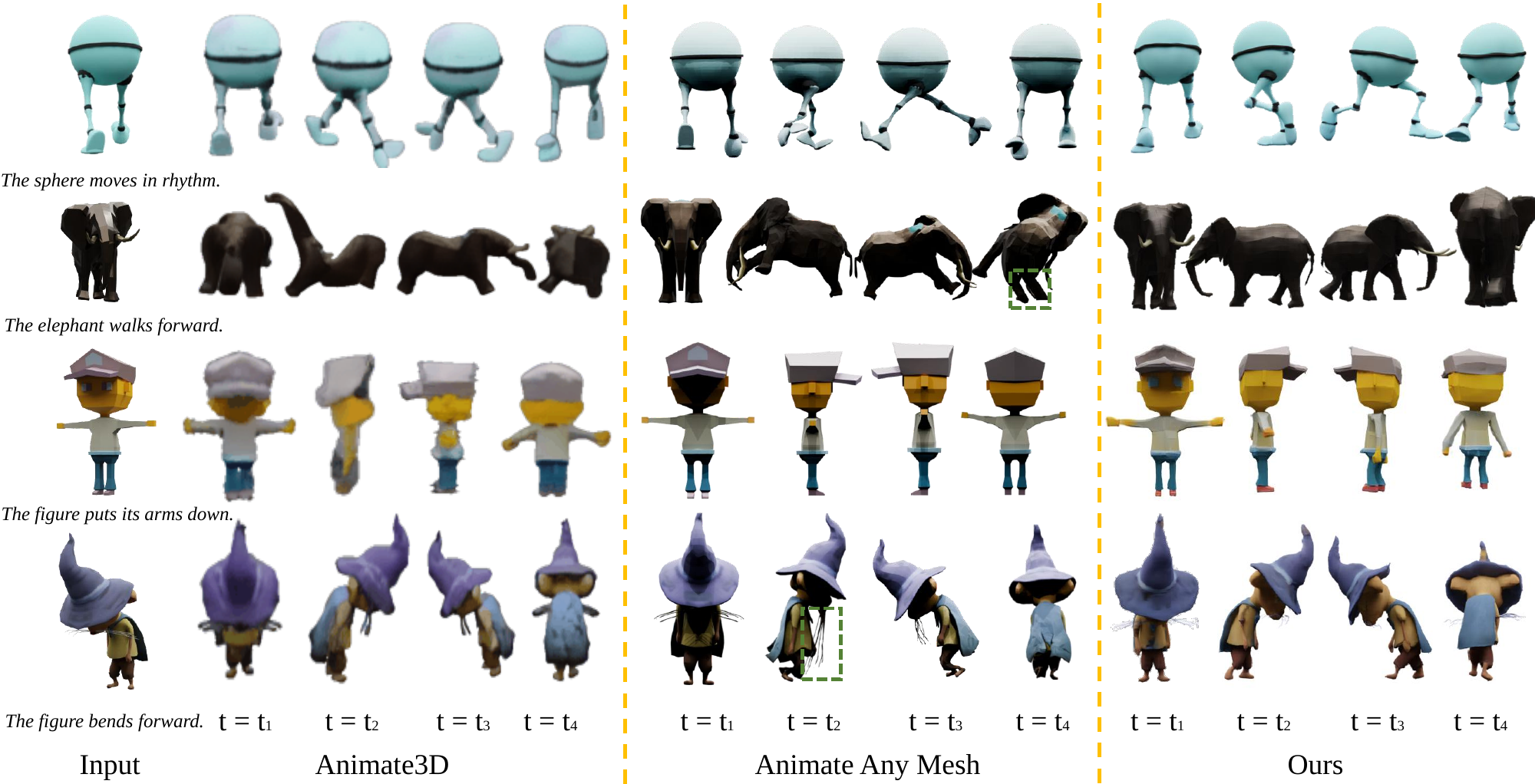}
    \vspace{-0.6cm}
    \caption{Qualitative comparison with baseline methods across diverse animation scenarios. TESLA produces smoother motions with better geometric stability. In contrast, AnimateAnyMesh shows limited deformation due to topology preservation challenges in vertex-based mesh animation, while Animate3D exhibits temporal flickering.}
    \label{fig:animation}
    \vspace{-0.4cm}
\end{figure}

\noindent\textbf{Baselines}. Following the data setting of~\cite{jiang2024animate3d,wu2025animateanymesh}, we collect 30 test cases that span different object categories and motion complexities for evaluation. We compare TESLA against two representative animation generation methods using a reference image and text prompt as input. Since AnimateAnyMesh~\cite{wu2025animateanymesh} requires mesh input rather than images, we first generate a 3D mesh from the reference image using Trellis~\cite{xiang2024structured}, followed by mesh simplification~\cite{knodt2024single} to satisfy the vertex count constraints imposed by their architecture. The method then produces animated meshes through trajectory decomposition and topology aware encoding. Animate3D~\cite{jiang2024animate3d} similarly leverages Trellis generated 3DGS as initialization, employing a video diffusion model finetuned on 3D data to synthesize multi-view consistent dynamic sequences, subsequently reconstructing 4D representations through optimization.

\noindent\textbf{Comparison}. As shown in Tab.~\ref{tab:animation}, TESLA consistently outperforms all baselines across both objective metrics and subjective evaluations. Unlike mesh-based methods that must move vertices while preserving topology, our approach naturally supports flexible deformation without such constraints. This enables smoother and more realistic animations, particularly for complex articulated motions. Fig.~\ref{fig:animation} illustrates this advantage. TESLA produces smooth animations with stable geometry, while Animate3D suffers from temporal flickering and AnimateAnyMesh exhibits limited deformation. The walking elephant example shows TESLA generating realistic gait patterns, whereas baselines display artifacts or stiffness. Complementary motion-centric metrics, including motion stability, prompt-conditioned diversity, and inter-frame consistency, are reported in Appendix~\ref{sec:appendix_motion}.

\begin{figure}[t]
    \centering
    \includegraphics[width=\linewidth]{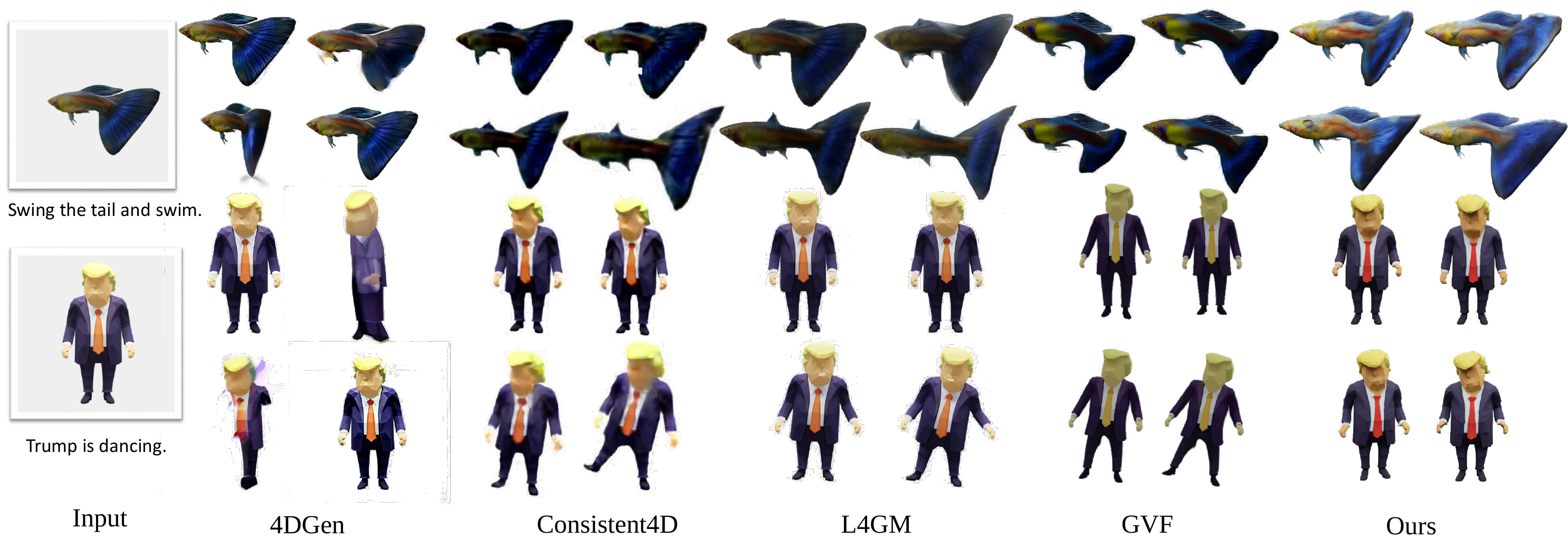}
    \vspace{-0.6cm}
    \caption{Qualitative comparison of 4DGS generation on Consistent4D dataset from text prompts across different methods.}
    \label{fig:consistent4d}
    \vspace{-0.4cm}
\end{figure}

\subsection{4DGS Generation}
\label{sec.4.2}
\noindent\textbf{Datasets and Metrics}. We follow the evaluation protocol from Consistent4D~\cite{jiang2024consistentd}. For evaluation, we adopt three complementary metrics: CLIP Score~\cite{radford2021learning} measures semantic alignment between rendered animations and text prompts, FVD~\cite{unterthiner2019fvd} evaluates the realism and quality of generated motion by comparing feature distributions with real sequences, and generation time quantifies computational efficiency on a single NVIDIA 4090 GPU.

\begin{wraptable}{R}{0.5\textwidth}
  \centering
  \vspace{-0.6cm}
  \setlength{\tabcolsep}{8pt}
  \caption{Quantitative comparison on 4DGS generation. * indicates methods that require video inputs, which we generate using Wan2.1~\cite{wan2025}.}
  \resizebox{0.5\textwidth}{!}{
    \begin{tabular}{l|ccc}
      \toprule
      Method & CLIP~$\uparrow$ & FVD~$\downarrow$ & Time~$\downarrow$  \\
      \midrule
      4DGen~\cite{yin20234dgen}                & 25.06 & 1713.2 & $\sim$1.5h \\
      Consistent4D*~\cite{jiang2024consistentd}            &  25.50  & 992.49   & $\sim$1.6h  \\
      L4GM*~\cite{ren2024l4gm}                & 25.72 &  \underline{876.25} &  \textbf{4.5s} \\
      GVF*~\cite{zhang2025gaussian}                & \underline{25.75} & 891.38  & 15.5s  \\
      Ours                 & \textbf{25.98} & \textbf{778.43} & \underline{9.2s} \\
      \bottomrule
    \end{tabular}
  }
  \label{tab:consistent4d}
  \vspace{-0.4cm}
\end{wraptable}

\noindent\textbf{Baselines}. We compare TESLA against four state-of-the-art 4D generation methods: 4DGen~\cite{yin20234dgen}, Consistent4D~\cite{jiang2024consistentd}, L4GM~\cite{ren2024l4gm}, and GVF~\cite{zhang2025gaussian}. Since TESLA natively accepts image and text inputs, we can directly evaluate 4DGen using the same input modality. However, the remaining baselines require video inputs for 4D reconstruction. To ensure fair comparison, we first generate multi-view video sequences from the reference image and text prompt using Wan2.1~\cite{wan2025}, a state-of-the-art video diffusion model. These generated videos are then fed into Consistent4D, L4GM, and GVF to produce their corresponding 4D representations.

\noindent\textbf{Comparison}. As shown in Tab.~\ref{tab:consistent4d}, TESLA achieves superior performance across all metrics while maintaining competitive generation speed. Our method demonstrates better semantic alignment with text prompts and generates more realistic motion with enhanced temporal coherence. Moreover, TESLA achieves competitive generation speed comparable to feed-forward methods while substantially faster than optimization-based approaches~\cite{yin20234dgen,jiang2024consistentd}. Fig.~\ref{fig:consistent4d} illustrates qualitative comparisons. TESLA produces smoother animations with stable geometry, while baselines suffer from temporal flickering or geometric artifacts. Additional qualitative comparisons against state-of-the-art baselines are provided in the appendix (Fig.~\ref{fig:additional_comparison}).

\begin{figure}[t]
  \centering
  \begin{minipage}{0.51\linewidth}
    \centering
    \vspace{0.1cm}
    \resizebox{\linewidth}{!}{
      \setlength{\tabcolsep}{12pt}
      \begin{tabular}{l|ccc}
        \toprule
        Config. & PSNR $\uparrow$ & SSIM $\uparrow$ & LPIPS $\downarrow$ \\
        \midrule
        w/o $\mathcal{L}_{\text{mot}}$ & 29.41 & 0.966 & 0.0417 \\
        w/o $\mathcal{L}_{\text{vol}}$ & 30.17 & 0.972 & 0.0393 \\
        \textbf{Full} & \textbf{31.82} & \textbf{0.979} & \textbf{0.0285} \\
        \bottomrule
      \end{tabular}
    }
    \vspace{0.2cm}
    \captionof{table}{VAE Loss Function Components.}
    \label{tab:vae_loss}
  \end{minipage}
  \hfill
  \begin{minipage}{0.47\linewidth}
    \centering
    \includegraphics[width=\linewidth]{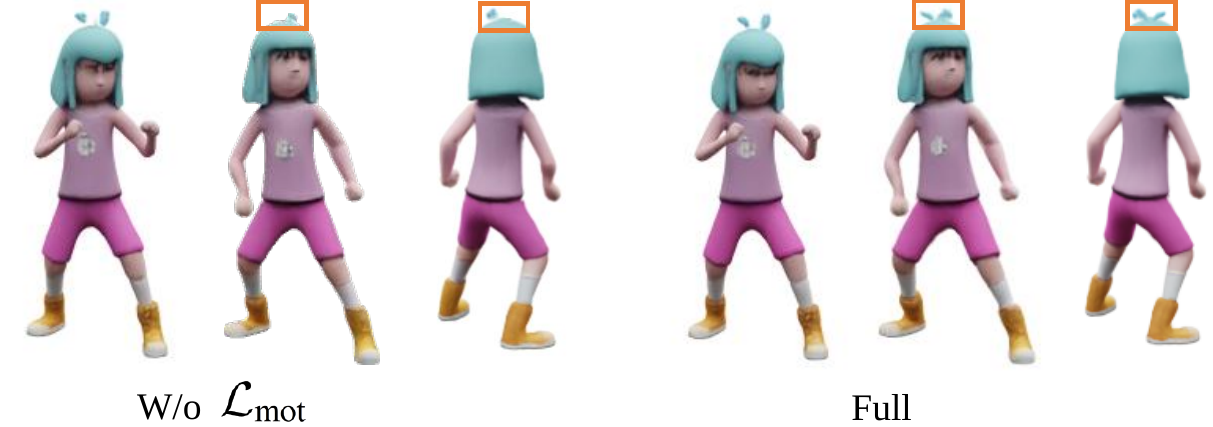}
    \vspace{-0.6cm}
    \caption{Visualizations on ablation study.}
    \label{fig:vae_ablation}
  \end{minipage}
  \vspace{-0.4cm}
\end{figure}

\subsection{Ablations and Further Analysis}
\label{sec.4.3}

\noindent\textbf{VAE Architecture Ablations}. We systematically evaluate the contribution of each regularization term in our VAE training objective to understand their individual and combined effects on reconstruction quality.
As demonstrated in Tab.~\ref{tab:vae_loss}, each component plays a distinct role in ensuring high-quality 4DGS generation. Specifically, motion regularization $\mathcal{L}_{\text{motion}}$ maintains temporal stability by directly constraining Gaussian velocities, preventing erratic movements across frames. Without this constraint, individual Gaussians may exhibit unpredictable motion patterns that break temporal coherence and lead to visual artifacts. Volume regularization $\mathcal{L}_{\text{vol}}$ prevents excessive expansion of Gaussian primitives by constraining their scales, thereby maintaining reasonable sizes throughout the animation sequence. Fig.~\ref{fig:vae_ablation} visualizes these effects: without $\mathcal{L}_{\text{motion}}$, the generated animations exhibit significant temporal jittering and inconsistent frame-to-frame transitions, whereas our full model produces smooth, coherent animations with stable geometric structures.

\begin{wrapfigure}[16]{r}{0.5\linewidth}
    \centering
    \vspace{-0.4cm}
    \includegraphics[width=\linewidth]{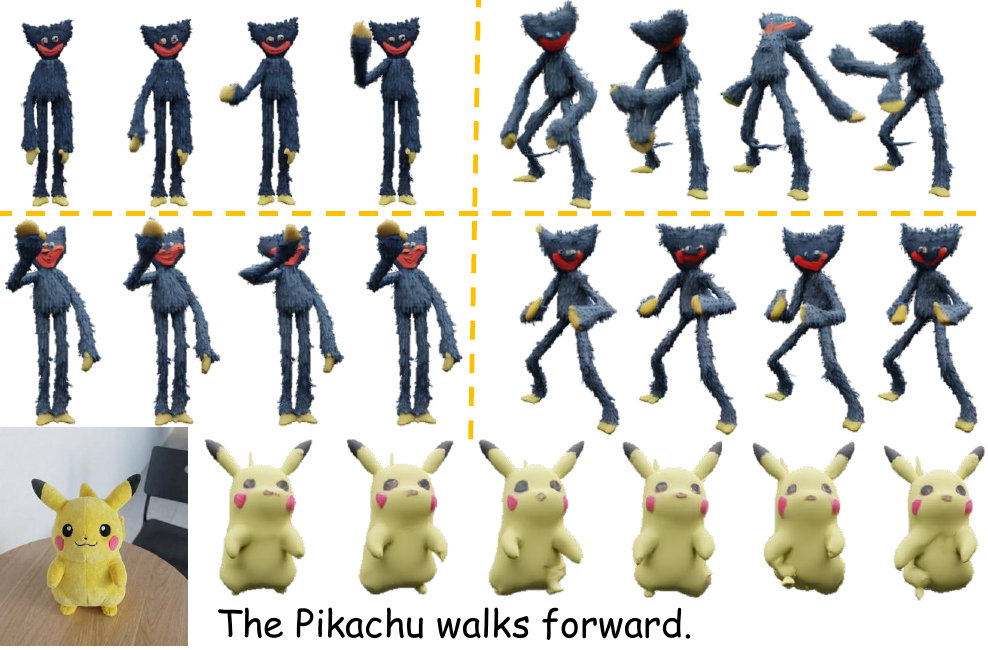}
    \vspace{-0.35cm}
    \caption{Motion diversity and in-the-wild generalization.}
    \label{fig:diversity_wild}
\end{wrapfigure}
\noindent\textbf{Generalization}. To evaluate TESLA's generalization beyond the curated training distribution, we examine two complementary aspects: motion diversity and in-the-wild applicability. Fig.~\ref{fig:diversity_wild} visualizes multi-motion generation from the same reference image conditioned on different text prompts, alongside test cases drawn from real-world sources outside the training data. TESLA produces temporally coherent and semantically aligned animations across both settings, demonstrating that the temporally structured latent space captures a rich distribution of motion patterns rather than collapsing to repetitive behaviors. We further quantify motion diversity in Appendix~\ref{sec:appendix_motion} to verify that TESLA learns diverse motions despite the limited scale of available 4D animation data. These results suggest that the decoupled geometry and appearance representation has internalized generalizable spatiotemporal priors that extend well beyond the training distribution.

\section{Conclusion}
In this work, we present TESLA, a novel feedforward approach for native 4D Gaussian Splatting generation. We leverage a diffusion model to learn distributions of temporally structured latent representations that decouple geometry evolution from appearance dynamics, enabling effective spatio-temporal modeling. Our native 4DGS decoder directly transforms temporal latents into Gaussian primitives. Evaluations on animation generation and 4DGS synthesis benchmarks demonstrate our superior performance over state-of-the-art methods in generating high-quality dynamic content with enhanced temporal coherence. We discuss limitations and broader impact in Appendix~\ref{sec:appendix_limitations}.

\begin{ack}
This work was partially supported by Deep Earth Probe and Mineral Resources Exploration -- National Science and Technology Major Project (2024ZD1003405) and the National Natural Science Foundation of China (62272263), and in part by Kuaishou.
\end{ack}

\bibliographystyle{ieee_fullname}
\bibliography{sample}

\newpage
\appendix
\section{More Implementation Details}
\subsection{Training Data Preparation}
\label{sec:appendix_data}
To train TESLA effectively, we curate a comprehensive dataset of animated 4D sequences that provides diverse temporal dynamics and geometric variations. Our dataset construction begins with collecting animated meshes from three primary sources: Deformthings4D~\cite{li20214dcomplete}, Objaverse~\cite{objaverse}, and Objaverse-XL~\cite{objaverseXL}. We filter sequences based on animation quality and temporal consistency, resulting in 45,000 animated sequences spanning various object categories.

\begin{figure}[t]
    \centering
    \includegraphics[width=0.85\linewidth]{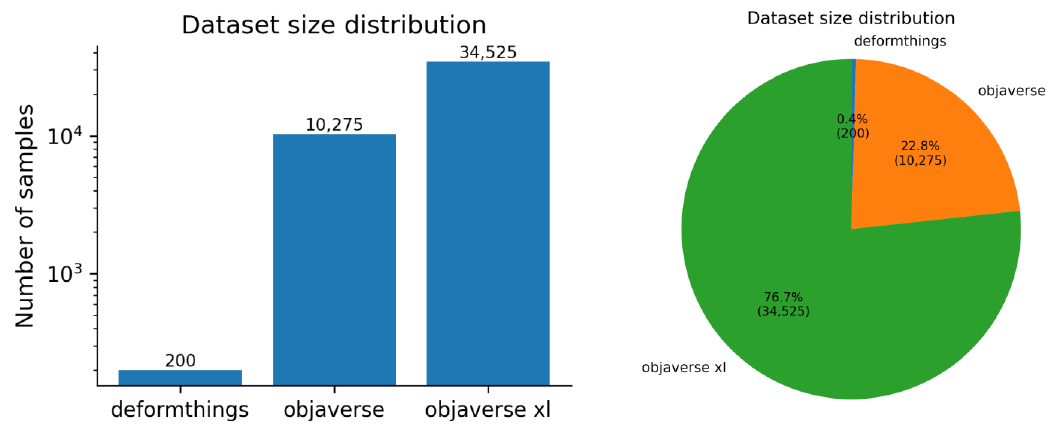}
    \vspace{-0.4cm}
    \caption{Representative samples from our dataset across various object categories.}
    \label{fig:dataset}
    \vspace{-0.4cm}
\end{figure}

For each filtered sequence, we sample 24 uniformly distributed frames to capture complete temporal evolution while maintaining computational feasibility. At each temporal sample, we render the object from 48 or 36 distinct camera viewpoints. Text descriptions are generated using Qwen2.5-VL~\cite{Qwen2.5-VL} to caption the rendered animation sequences, providing accurate semantic descriptions of temporal dynamics and motion patterns. This comprehensive rendering pipeline requires approximately 50,000 GPU hours, resulting in a dataset where each training sample consists of reference images, text descriptions, and multi-view rendered frames across the temporal sequence. Fig.~\ref{fig:dataset} illustrates representative examples from our curated dataset, showcasing the diversity of object categories and motion patterns.

\subsection{Temporal Latent Encoding}
\label{sec:appendix_encoding}
We directly inherit the encoder of TRELLIS~\cite{xiang2024structured} without modification and keep it frozen throughout training. Since the ground-truth sequences are temporally consistent and free of temporal artifacts, we apply this encoder independently at each timestep. At timestep $t$, we voxelize the frame onto the same $R^3$ grid used by the latent representation, which gives the active voxel positions $\{p_{i,t}\}$, and render the frame from the 36 or 48 viewpoints described in Sec.~\ref{sec:appendix_data}. DINOv2~\cite{oquab2024dinov2} features extracted from each rendered view are projected onto the active voxels and averaged across views, and a sparse VAE encoder then compresses the resulting voxel features into the latents $\{z_{i,t}\}$. Repeating this procedure for $t = 1, \ldots, T$ yields the target latent sequence $S = \{S_t\}_{t=1}^{T}$, which supervises the decoder pre-training and both flow models. Only the 4DGS decoder, the SS flow model, and the slat flow model are optimized (Tab.~\ref{tab:model}).

Because the encoder is inherited, we make no novelty claim about its design. Encoding frames independently still preserves sufficient information, since decoding the encoded ground-truth latents reaches a PSNR of 31.82, an SSIM of 0.979, and an LPIPS of 0.0285 (Tab.~\ref{tab:vae_loss}). As the encoder has no temporal mechanism of its own, the temporal coherence of generated sequences must come from the downstream models, and removing temporal attention indeed lowers I2V by 0.049 (Tab.~\ref{tab:architecture_ablation}).

Our 4DGS decoder, unlike the encoder, differs substantially from its TRELLIS counterpart. TRELLIS decodes a static latent into 3D Gaussians, whereas our decoder maps the temporal latent sequence into 4D Gaussians with 4D centers, 4D scales, isoclinic quaternion pairs, and a learned temporal offset $\Delta t$, so the output can be rendered at arbitrary continuous timesteps rather than only at discrete frames. Each active voxel is decoded into $K$ Gaussians, indexed by $k$ in Eqs.~(\ref{eq:vol}) and~(\ref{eq:alpha}), whose 4D centers are predicted in continuous space.

\subsection{Training Details}
We train TESLA in three stages following the pipeline described in the main paper. In all stages, we use AdamW as the optimizer with a learning rate of \(1 \times 10^{-4}\), \(\beta_1 = 0.9\), \(\beta_2 = 0.999\), and weight decay of \(0.01\). The learning rate is scheduled with a cosine decay and a short linear warm-up at the beginning of training. Unless otherwise noted, we sample sequences of 24 frames and render each frame from 36 or 48 views, forming mini-batches over multiple sequences and views to fully utilize GPU memory. We adopt mixed-precision training and gradient clipping to stabilize optimization. The overall training objective is a weighted sum of per-pixel reconstruction and regularization losses: we combine an \(\ell_1\) loss with SSIM and LPIPS losses for image reconstruction, and add motion, volume, and opacity regularization terms to encourage temporally smooth trajectories and compact, non-degenerate Gaussian sets, using \(\lambda_{\text{motion}} = 0.1\), \(\lambda_{\text{vol}} = 10{,}000.0\), and \(\lambda_{\alpha} = 0.1\) in all experiments. The 4DGS decoder is first pre-trained on ground-truth multi-view supervision, after which we train the rectified flow model for sparse temporal structure prediction (SS flow), and finally the rectified flow model for latent prediction (slat flow). All large-scale experiments are conducted on NVIDIA A800 GPUs; training the full TESLA model requires several days on 32 GPUs, while the decoder pre-training stage is significantly cheaper and can be completed on a single node.

\subsection{Model Architecture}
Our model consists of three main components: the 4DGS decoder, the temporal SS flow model, and the temporal slat flow model. The 4DGS decoder employs a transformer-based architecture with temporal attention mechanisms to directly predict Gaussian parameters from structured latents. Both flow models utilize Diffusion Transformer architectures with parallel attention mechanisms for multi-modal conditioning. Tab.~\ref{tab:model} provides detailed architectural specifications for each component.

\subsection{Inference Pipeline}
\label{sec:appendix_inference}
During inference, TESLA generates native 4D Gaussian Splatting representations through a two-stage process. Given a reference image $I$ and text prompt $P$, we encode them into conditioning features and initialize with random noise. Our rectified flow model first generates sparse temporal voxel structures by predicting positional indices $p_{i,t} \in \{0,1,...,(N-1)\}^3$ for active voxels at each timestep $t$, establishing the geometric scaffold. Conditioned on this structure, the model then synthesizes latent features $z_{i,t} \in \mathbb{R}^C$ for each active voxel position, completing the temporally structured latent representation $S = \{S_t\}_{t=1}^T$ where $S_t = \{(z_{i,t}, p_{i,t})\}_{i=1}^{L_t}$. The latents are directly transformed into native 4D Gaussian primitives through the 4DGS decoder, enabling flexible rendering at arbitrary viewpoints and timesteps.

\section{Evaluation Metrics}
We evaluate TESLA on both text and image-conditioned 4D generation tasks and compare it against a set of strong baselines under a unified evaluation protocol. Unless otherwise specified, all methods generate sequences of 24 frames with shared camera trajectories.

\subsection{Quantitative metrics}
We use a combination of distributional, semantic, and low-level metrics. For distributional similarity, we compute the Fréchet Video Distance (FVD) between generated videos \(\mathcal{G}\) and real videos \(\mathcal{R}\). Let \((\mu_g, \Sigma_g)\) and \((\mu_r, \Sigma_r)\) be the mean and covariance of feature embeddings extracted from a pretrained video recognition network for \(\mathcal{G}\) and \(\mathcal{R}\), respectively. FVD is given by
\[
\mathrm{FVD}(\mathcal{G}, \mathcal{R})
= \|\mu_g - \mu_r\|_2^2
+ \mathrm{Tr}\Big( \Sigma_g + \Sigma_r - 2 (\Sigma_g^{1/2} \Sigma_r \Sigma_g^{1/2})^{1/2} \Big),
\]
where lower values indicate better alignment with the real data distribution. To measure semantic alignment for text-conditioned generation, we compute CLIP-based similarities. Let \(f_{\text{text}}(x_i)\) be the CLIP text embedding and \(f_{\text{img}}(\hat{I}^{M}_{i,t,v})\) the image embedding of a generated frame. We calculate
\[
s_{\text{text}}^{M}(i)
= \frac{1}{TV} \sum_{t=1}^T \sum_{v=1}^V
\frac{ \langle f_{\text{text}}(x_i),\, f_{\text{img}}(\hat{I}^{M}_{i,t,v}) \rangle }
     { \| f_{\text{text}}(x_i) \|_2 \, \| f_{\text{img}}(\hat{I}^{M}_{i,t,v}) \|_2 },
\]
 and report the average \(\bar{s}_{\text{text}}^{M} = \frac{1}{N}\sum_{i=1}^N s_{\text{text}}^{M}(i)\). For image-conditioned generation, we instead compare frames to the reference image \(x_i^{\text{ref}}\) using the same formulation.

\subsection{VBench evaluation}
To obtain fine-grained assessments of motion quality and temporal behavior, we additionally employ VBench~\cite{zhang2024evaluationagent}. For each generated sequence, we render the motion as a 24-frame video and feed the full sequence into the VBench pipeline using the default configuration. VBench produces multiple dimension-wise scores such as subject fidelity, motion smoothness, temporal consistency, and aesthetic quality. We evaluate all methods under identical settings and report the averaged score over all prompts and sequences for each dimension.

\subsection{Baseline implementations}

For animation generation, we follow the evaluation protocol of Animate3D~\cite{jiang2024animate3d} and AnimateAnyMesh~\cite{wu2025animateanymesh}. Since AnimateAnyMesh operates on meshes, we first reconstruct 3D meshes from reference images using Trellis~\cite{xiang2024structured} and apply mesh simplification~\cite{knodt2024single} to meet vertex-count constraints. Similarly, Animate3D initializes with Trellis-generated 3DGS and employs a video diffusion model fine-tuned on 3D data to synthesize multi-view dynamic sequences for 4D reconstruction.

For 4DGS generation, we follow Consistent4D~\cite{jiang2024consistentd}'s protocol and compare against four state-of-the-art methods: 4DGen~\cite{yin20234dgen}, Consistent4D~\cite{jiang2024consistentd}, L4GM~\cite{ren2024l4gm}, and GVF~\cite{zhang2025gaussian}. TESLA and 4DGen directly accept image-text inputs for evaluation. Since Consistent4D, L4GM, and GVF require video inputs, we first generate multi-view video sequences from the reference images and text prompts using Wan2.1~\cite{wan2025}, then feed these videos into their official implementations. All methods are evaluated on a single NVIDIA 4090 GPU using CLIP score, FVD, and generation time to ensure consistent comparison of quality and efficiency.

For these video inputs, we use Wan2.1-I2V-14B-720P~\cite{wan2025}. We choose the image-to-video variant rather than the text-to-video variant because it conditions on the same reference image that every method receives, which keeps the inputs aligned across the comparison. We keep the official default inference settings, namely a classifier-free guidance scale of 5.0, a flow-matching shift of 5.0, a maximum resolution of $1280\times720$ with the aspect ratio determined by the reference image, the default negative prompt of the I2V configuration, and disabled prompt extension. Each generated video contains 81 frames at 16 fps, from which we uniformly sample the 24 frames used in our evaluation. The same generated videos are given to all three video-conditioned baselines, whereas TESLA receives only the reference image and the text prompt and never accesses these videos.

\begin{table*}[t]
\centering
\caption{More details about Model Architectures.}
\label{tab:model}
\resizebox{\textwidth}{!}{
\begin{tabular}{l|c|c|c|l|l|c}
\toprule
Network & \#Layer & \#Dim. & \#Head & Block Arch. & Special Modules & \#Param. \\
\midrule
$\mathcal{D}_{4DGS}$ & 12 & 768 & 12 & SW-MSA + FFN & Swin Attn. & 85.52M \\
SS Flow & 24 & 1024 & 16 & Parallel Attn + Temporal Attn & QK Norm. +  Sp. Conv. & 643.84M \\
Slat Flow & 24 & 1024 & 16 & Parallel Attn + Temporal Attn & QK Norm. +  Sp. Conv. & 688.61M \\
\bottomrule
\end{tabular}
}
\vspace{-0.3cm}
\end{table*}

\begin{table}[t]
\centering
\caption{Ablation studies on key architectural components.}
\vspace{-0.2cm}
\label{tab:architecture_ablation}
\setlength{\tabcolsep}{20pt}
\begin{tabular}{l|ccc}
\toprule
Variant & I2V~$\uparrow$ & M.Sm.~$\uparrow$ & Sub.Cons.~$\uparrow$ \\
\midrule
w/o Temporal Attn & 0.812 & 0.991 & 0.914 \\
Sequential Attn & 0.851 & 0.992 & 0.927 \\
Spatial only (xyz) & 0.811 & 0.986 & 0.907 \\
\textbf{Full} & \textbf{0.861} & \textbf{0.994} & \textbf{0.941}  \\
\bottomrule
\end{tabular}
\vspace{-0.3cm}
\end{table}

\section{More Experiments}
\subsection{Additional Ablation Studies}
We conduct additional ablation studies to validate the effectiveness of our key architectural designs. Table~\ref{tab:architecture_ablation} presents a comprehensive analysis of three critical components: temporal attention mechanism, parallel attention architecture, and position encoding strategy.

The results demonstrate that temporal attention plays a crucial role in capturing motion dynamics, as removing it leads to significant performance degradation. For multi-modal conditioning, our parallel architecture consistently outperforms sequential fusion alternatives, validating its effectiveness in integrating image and text guidance. Finally, we find that using only spatial or temporal encoding alone yields degraded performance, while our hybrid spatio-temporal encoding explicitly disentangles these two dimensions, enabling the network to better distinguish and model geometric structure and motion dynamics separately for coherent 4D generation.

\subsection{Geometric Accuracy}
\begin{wraptable}{r}{0.32\linewidth}
    \centering
    \vspace{-0.6cm}
    \caption{Geometric Accuracy.}
    \label{tab:uni3d}
    \setlength{\tabcolsep}{8pt}
    \begin{tabular}{l|c}
        \toprule
         & Uni3D $\uparrow$    \\
        \midrule
        4DGen~\cite{yin20234dgen}             &  24.25 \\
        C4D~\cite{jiang2024consistentd}       & 10.55  \\
        L4GM~\cite{ren2024l4gm}               &  33.81 \\
        GVF~\cite{zhang2025gaussian}                &  35.66 \\
        Ours               &  \textbf{35.89} \\
        \bottomrule
    \end{tabular}
    \vspace{-0.4cm}
\end{wraptable}
The latent grid resolution in TESLA is carefully designed to strike a balance between representation efficiency and geometric fidelity. The chosen resolution demonstrates sufficient capacity to handle geometrically challenging structures while maintaining practical efficiency. To further substantiate this with quantitative evidence, we evaluate geometric accuracy using the Uni3D metric in Tab.~\ref{tab:uni3d}, where TESLA achieves competitive performance against prior methods and confirms that the chosen grid resolution does not compromise structural precision.

\subsection{Variable Length}
Although TESLA is trained on fixed-length sequences, the temporally structured latent representation is inherently capable of accommodating different sequence lengths without architectural modifications. To validate this flexibility, we conduct an ablation study comparing models trained on sequences of different lengths. As reported in Tab.~\ref{tab:data_2}, models trained on shorter sequences achieve comparable performance across all evaluation metrics, confirming that TESLA generalizes well to variable temporal granularities and does not rely on a fixed sequence length to maintain generation quality.
\begin{table}[t]
    \centering
    \caption{Ablation studies on sequence length.}
    \vspace{-0.2cm}
    \setlength{\tabcolsep}{13pt}
    \begin{tabular}{l|cccc}
        \toprule
          & I2V~$\uparrow$ & M.Sm.~$\uparrow$ & Sub.Cons.~$\uparrow$ & Aest.Q~$\uparrow$  \\
        \midrule
        16 frames & 0.842 & \textbf{0.994} &  0.922 & \textbf{0.534} \\
        24 frames & \textbf{0.861} &  \textbf{0.994} & \textbf{0.941} & 0.533 \\
        \bottomrule
    \end{tabular}
    \label{tab:data_2}
    \vspace{-0.3cm}
\end{table}

\begin{figure}[t]
    \centering
    \includegraphics[width=\linewidth]{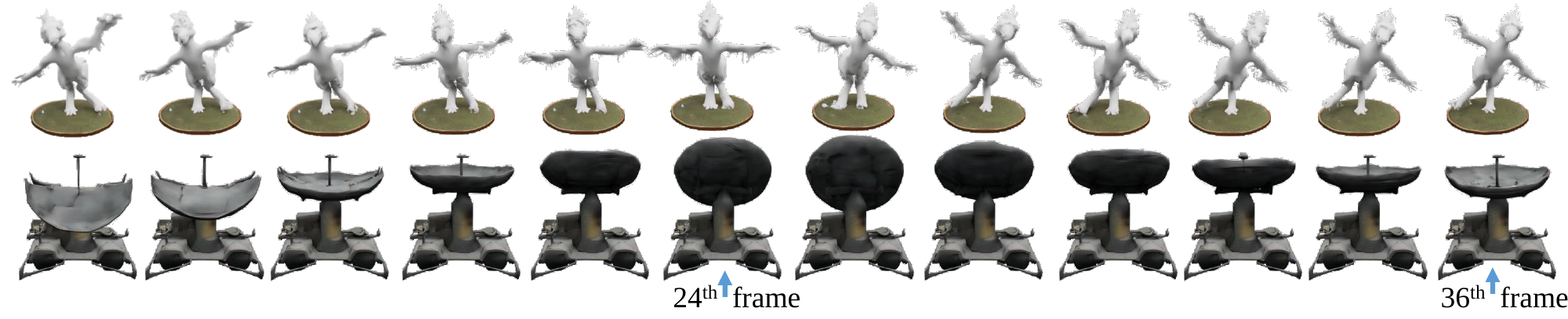}
    \vspace{-0.6cm}
    \caption{TESLA generalizes to extended-length sequences via sliding window inference.}
    \label{fig:more}
    \vspace{-0.3cm}
\end{figure}

To extend generation beyond the training-time horizon, we adopt a sliding window strategy at inference, drawing from inpainting techniques widely used in diffusion models. Starting from an initial generation, the final half of the generated frames serve as inpainting conditions for the subsequent window, establishing a temporal overlap that guarantees continuity across window boundaries. Through this iterative process, sequences of arbitrary length can be produced while preserving consistent geometry and appearance. As demonstrated in Fig.~\ref{fig:more}, sequences generated through this approach exhibit smooth and coherent animations throughout window transitions, further validating TESLA's capability to produce long-horizon dynamic content that remains temporally stable beyond the training-time sequence length.

\needspace{24\baselineskip}
\subsection{Additional Qualitative Comparisons}
\label{sec:appendix_more_compare}
\leavevmode\par\vspace{-\baselineskip}
\begin{wrapfigure}[21]{r}{0.46\linewidth}
    \centering
    \vspace{-0.95cm}
    \includegraphics[width=\linewidth]{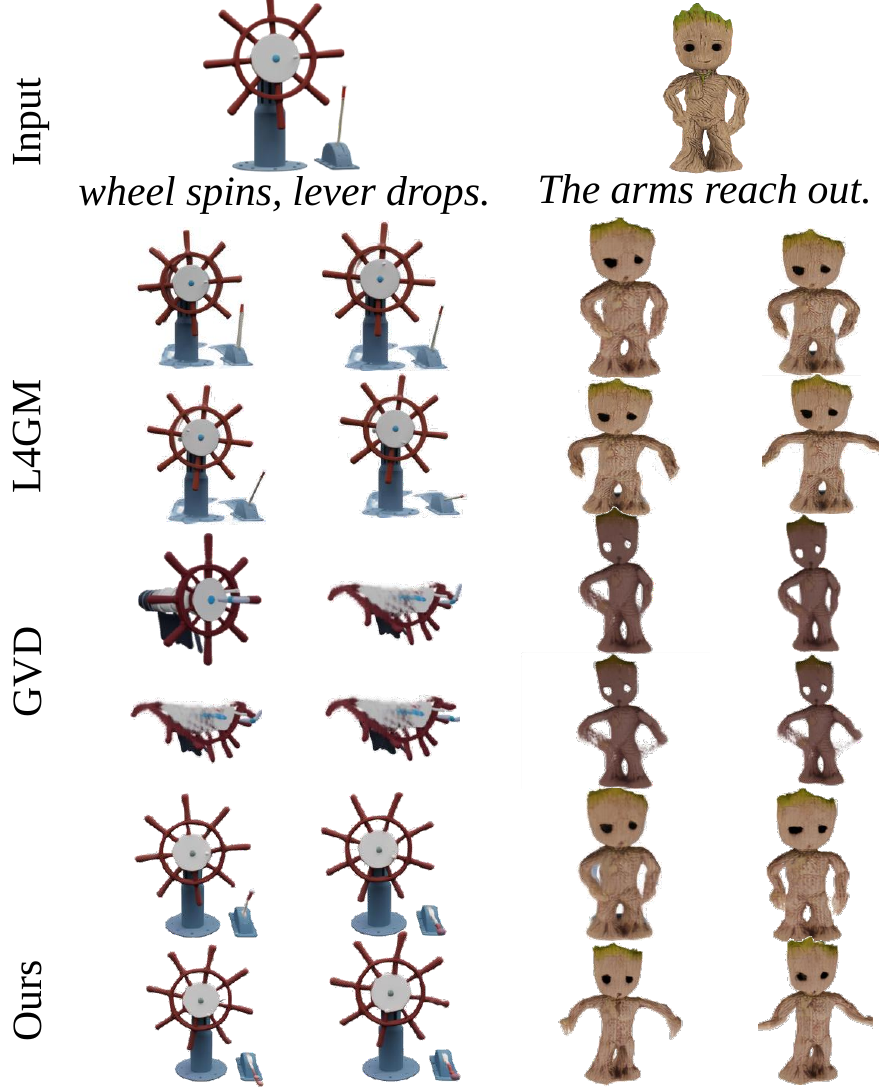}
    \vspace{-0.65cm}
    \caption{Additional qualitative comparisons on 4DGS generation.}
    \label{fig:additional_comparison}
    \vspace{-0.85cm}
\end{wrapfigure}
We provide additional qualitative comparisons against state-of-the-art 4DGS generation baselines in Fig.~\ref{fig:additional_comparison}, covering diverse object categories and motion patterns drawn from the Consistent4D~\cite{jiang2024consistentd} evaluation protocol. Across these cases, optimization-based approaches such as 4DGen~\cite{yin20234dgen} and Consistent4D~\cite{jiang2024consistentd} tend to produce over-smoothed surfaces and noticeable temporal flickering, while feed-forward baselines L4GM~\cite{ren2024l4gm} and GVF~\cite{zhang2025gaussian} inherit residual artifacts from the intermediate multi-view video synthesis stage and exhibit blurred appearance under large motion. In contrast, by directly generating 4D Gaussian primitives from temporally structured latents, TESLA bypasses these intermediate failure modes and yields more stable geometry, sharper appearance details, and smoother motion across the full sequence. These observations are consistent with the quantitative results reported in Tab.~\ref{tab:consistent4d}, further supporting the effectiveness of native 4DGS generation.

\subsection{Motion Quality and Diversity}
\label{sec:appendix_motion}
Beyond the view-aligned VBench scores reported in the main paper, we further evaluate three motion-centric properties that frame-level perceptual metrics do not directly capture. Dynamic Score is computed as the mean optical flow magnitude between consecutive rendered frames, where lower values indicate smoother temporal evolution with fewer flickering or jittering artifacts. Motion Diversity follows the Average Pairwise Distance protocol of~\cite{curreli2025nonisotropic}: for each reference object we render multiple animations conditioned on different text prompts and report the mean pairwise CLIP-feature distance, where larger values reflect richer prompt-conditioned motion variability. Temporal Consistency is computed using the VBench evaluation pipeline~\cite{zhang2024evaluationagent} based on pretrained vision-feature similarity between consecutive rendered frames. As summarized in Tab.~\ref{table:motion}, TESLA attains the lowest Dynamic Score and the highest Motion Diversity and Temporal Consistency, confirming that the temporally structured latent simultaneously suppresses unstable motion artifacts and admits diverse, semantically aligned animations. Qualitative evidence of this controllability is shown in Fig.~\ref{fig:diversity_wild}, where the same reference object is animated under multiple text prompts to elicit distinct motion patterns.

\begin{table}[h!]
    \centering
    \small
    \caption{Quantitative motion evaluation against animation baselines.}
    \vspace{-0.2cm}
    \setlength{\tabcolsep}{1.5pt}
    \resizebox{\linewidth}{!}{
        \begin{tabular}{l|ccc}
        \toprule
        Method & Dynamic Score $\downarrow$ & Motion Diversity $\uparrow$ & Temporal Consistency $\uparrow$ \\ \midrule
        Animate3D~\cite{jiang2024animate3d}        & 4.37          & 0.48          & 0.980 \\
        AnimateAnyMesh~\cite{wu2025animateanymesh} & 3.72          & 0.73          & 0.989 \\
        Ours                                       & \textbf{2.35} & \textbf{0.74} & \textbf{0.991} \\
        \bottomrule
        \end{tabular}
    }
    \label{table:motion}
    \vspace{-0.3cm}
\end{table}

\subsection{Comparison with Per-Frame TRELLIS}
\label{sec:appendix_trellis}
TESLA builds on the two-stage structure of TRELLIS~\cite{xiang2024structured}, which raises the question of whether native 4D generation is needed at all. To answer it, we evaluate TRELLIS independently on each frame using the same inputs and stack the resulting per-frame outputs. As shown in Tab.~\ref{tab:trellis}, per-frame TRELLIS performs worse on every temporal metric. This gap is structural. Independent per-frame generation establishes no cross-frame correspondence, and no Gaussian primitive persists over time, which leads to flickering across timesteps. A stack of independently generated 3DGS can also be rendered only at the discrete timesteps for which it was decoded, whereas TESLA renders at arbitrary continuous timesteps. TESLA avoids both problems because it jointly models the complete 4D structure of the sequence through temporal attention and the $xyzt$ position encoding, rather than generating sparse structures frame by frame.

\begin{table}[t]
    \centering
    \caption{Comparison with per-frame TRELLIS. Continuous $t$ indicates whether the output can be rendered at arbitrary continuous timesteps.}
    \vspace{-0.2cm}
    \label{tab:trellis}
    \setlength{\tabcolsep}{6pt}
    \begin{tabular}{l|cccc}
        \toprule
        Method & Dynamic Score~$\downarrow$ & Temp.Cons.~$\uparrow$ & Sub.Cons.~$\uparrow$ & Continuous $t$ \\
        \midrule
        Per-frame TRELLIS~\cite{xiang2024structured} & 4.210 & 0.979 & 0.938 & \xmark \\
        Ours & \textbf{2.350} & \textbf{0.991} & \textbf{0.941} & \cmark \\
        \bottomrule
    \end{tabular}
\end{table}

\subsection{More Results}
To further demonstrate the versatility and quality of TESLA, we provide additional generation results across diverse object categories and motion patterns in the following Fig~\ref{fig:frontview_1} and Fig~\ref{fig:free_1}.

\begin{figure*}[t]
    \centering
    \includegraphics[width=0.98\linewidth]{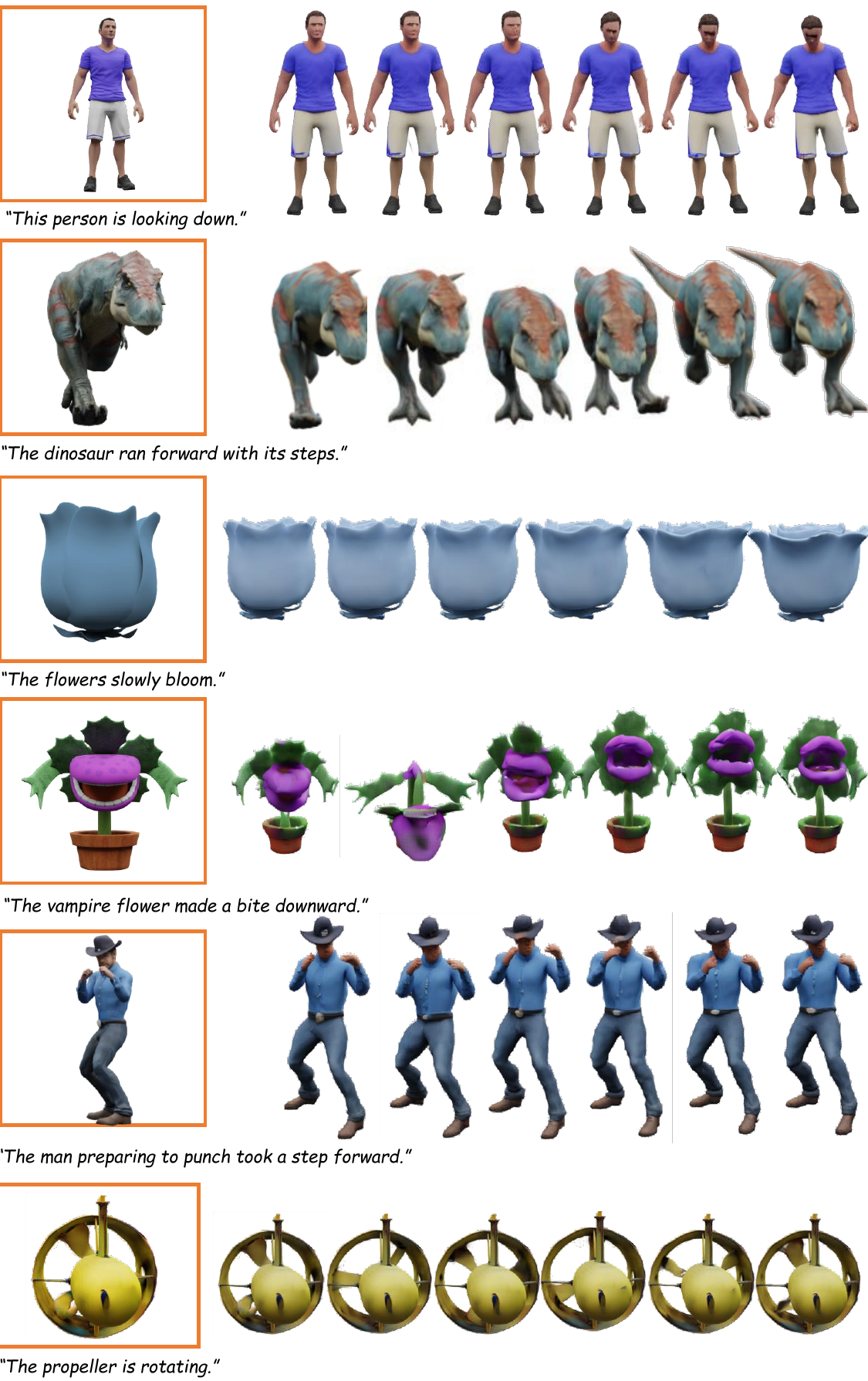}
    \vspace{-0.2cm}
    \caption{More visualizations of generated 4DGS results from front views (Part 1).}
    \label{fig:frontview_0}
\end{figure*}

\begin{figure*}[t]
    \centering
    \includegraphics[width=0.98\linewidth]{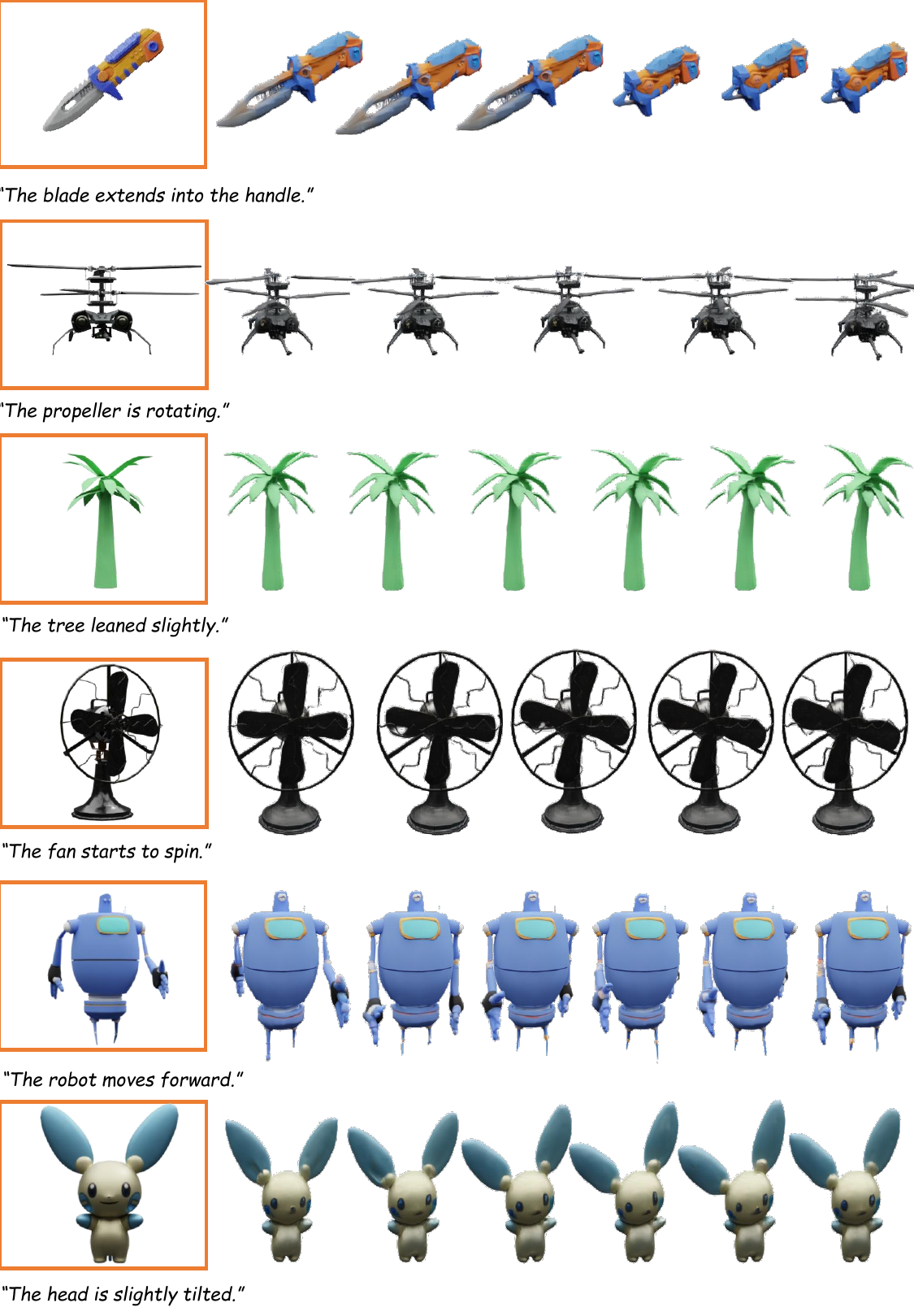}
    \vspace{-0.2cm}
    \caption{More visualizations of generated 4DGS results from front views (Part 2).}
    \label{fig:frontview_1}
\end{figure*}

\begin{figure*}[t]
    \centering
    \includegraphics[width=0.98\linewidth]{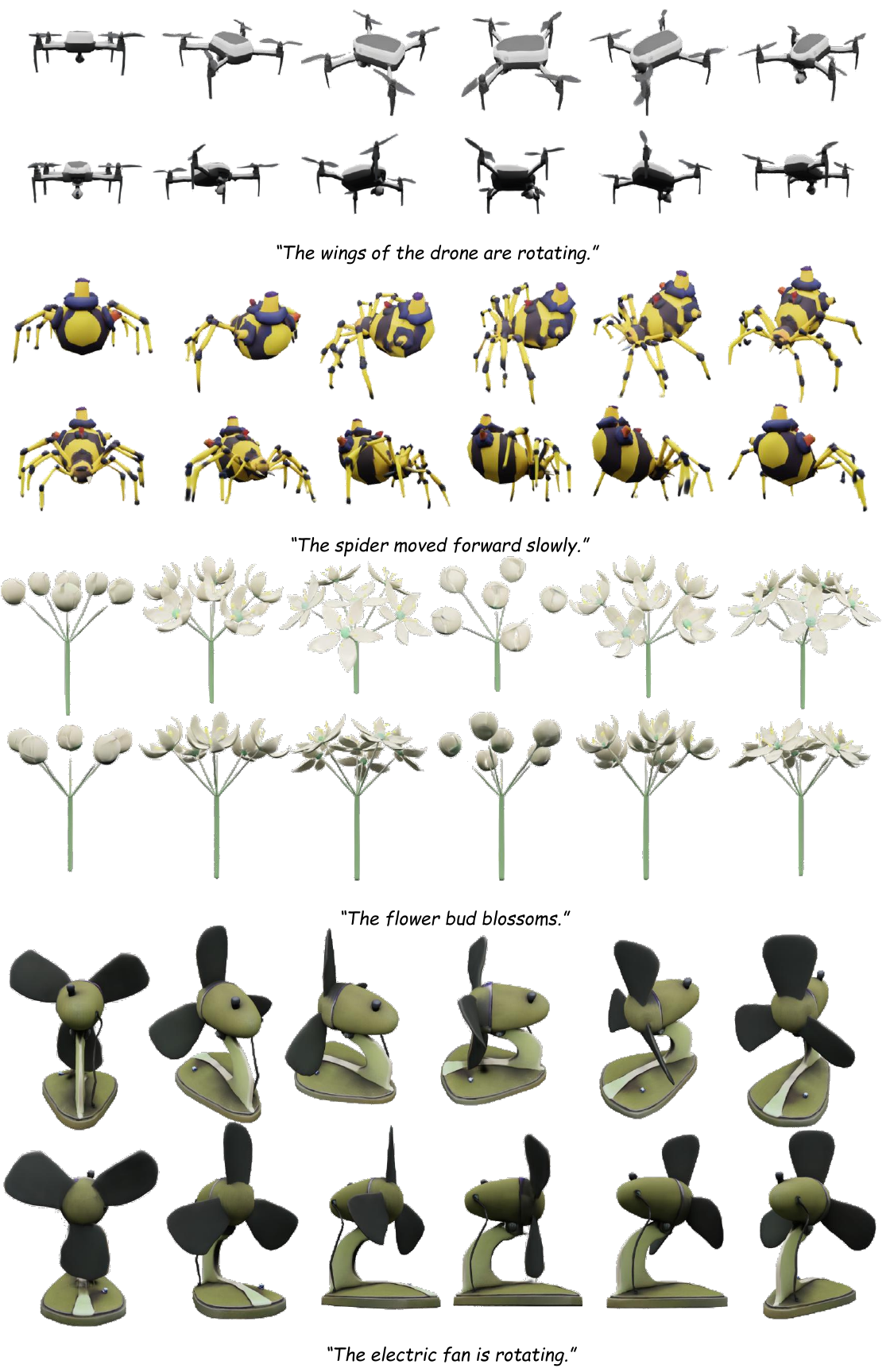}
    \vspace{-0.2cm}
    \caption{More visualizations of generated 4DGS results from free viewpoints (Part 1).}
    \label{fig:free_0}
\end{figure*}

\begin{figure*}[t]
    \centering
    \includegraphics[width=0.98\linewidth]{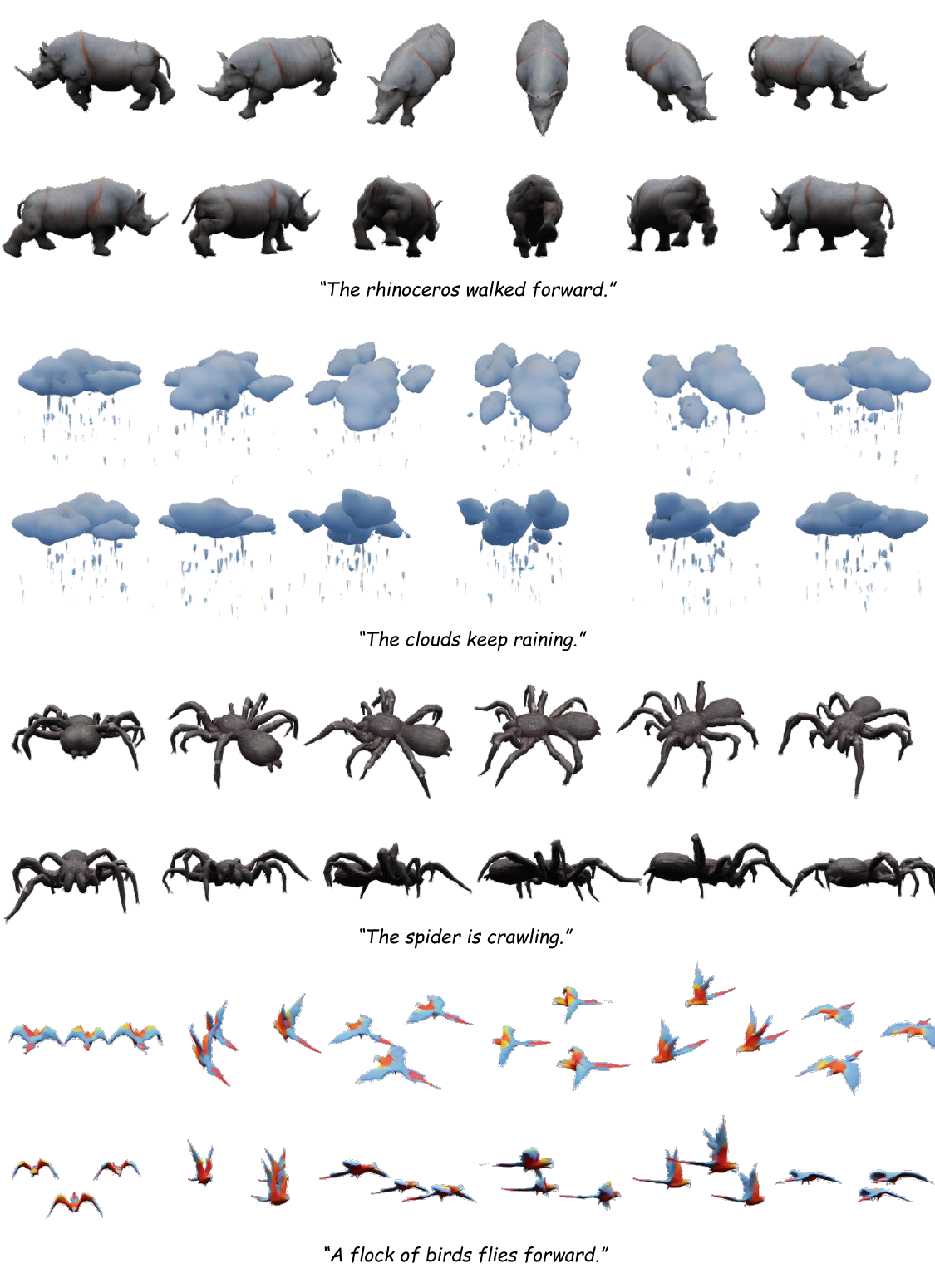}
    \vspace{-0.2cm}
    \caption{More visualizations of generated 4DGS results from free viewpoints (Part 2).}
    \label{fig:free_1}
\end{figure*}

\section{Concurrent Work}
\label{sec:appendix_concurrent}
Helix4D~\cite{yenphraphai2026helix4d}, which appeared after our submission, also extends the TRELLIS family to 4D. It adapts TRELLIS.2 from image-to-3D generation to video-conditioned 4D mesh generation, sharing information across frames through sliding-window cross-frame attention anchored on the first frame and injecting time through a 4D positional encoding that repurposes low-frequency spatial RoPE bands. The two methods address different settings. Helix4D is conditioned on an input video that specifies the motion, whereas TESLA generates the motion from a text prompt given only a reference image, and it outputs native 4D Gaussians rather than dynamic meshes.

\section{Limitations and Broader Impact}
\label{sec:appendix_limitations}
TESLA is currently evaluated on object-centric dynamic assets and finite-length animations, so performance on long multi-object interactions, complex scene-level dynamics, and out-of-distribution motion categories remains less explored.

Within this scope, we observe three main limitations. First, quality degrades as motion becomes larger or deformation more complex, since each 4D Gaussian has a limited temporal extent and cannot represent motion that spans much of the sequence. Second, detail is bounded by the sparse voxel lattice, so thin structures near the voxel resolution, such as animal limbs, may intermittently disappear, and fine textures such as faces can appear blurry, especially because high-fidelity faces are rare in our training data. Third, each 4D Gaussian evolves independently without a rigidity or physics prior. This flexibility enables topological changes, but rigid objects may not move rigidly and the motion may not be physically accurate, partly because most training sequences are artist-created. Higher lattice resolution, pretrained 2D diffusion priors, and physics-aware constraints are promising directions for future work.

As a generative 4D content system, TESLA may lower the cost of creating dynamic assets for AR/VR, games, and digital media, while also requiring responsible release practices to reduce misuse in deceptive or unauthorized synthetic content generation.

\end{document}